\documentclass[conf]{new-aiaa}
\usepackage[utf8]{inputenc}

\usepackage{graphicx}
\usepackage{amsmath}
\usepackage[version=4]{mhchem}
\usepackage{siunitx}
\usepackage{longtable,tabularx}
\usepackage{float}
\usepackage{caption}
\usepackage{subcaption}

\title{Taking a PEEK into YOLOv5 for Satellite Component Recognition via Entropy-based Visual Explanations\footnote{Article accepted at AIAA SCITECH Forum 2024\\ Copyright © 2024 by the American Institute of Aeronautics and Astronautics, Inc. All rights reserved.}}

\author{Mackenzie J. Meni\footnote{PhD Student, Department of Mathematics and Systems Engineering.}}
\affil{NEural TransmissionS Lab, Department of Mathematics and Systems Engineering \\ Florida Institute of Technology, Melbourne, FL 32901, USA}

\author{Trupti Mahendrakar\footnote{PhD Candidate, Department of Aerospace, Physics and Space Sciences, AIAA Student Member.}}
\affil{Autonomy Lab, Department of Aerospace, Physics and Space Sciences \\ Florida Institute of Technology, Melbourne, FL 32901, USA}

\author{Olivia D. M. Raney\footnote{MFA Student, Communication Media Arts, Scripps College of Communication.}}
\affil{Communication Media Arts, Scripps College of Communication,  \\ Ohio University, Athens, OH 45701, USA}

\author{Ryan T. White\footnote{Assistant Professor, Department of Mathematics and Systems Engineering.}}
\affil{NEural TransmissionS Lab, Department of Mathematics and Systems Engineering \\ Florida Institute of Technology, Melbourne, FL 32901, USA.}

\author{Michael L. Mayo\footnote{Research Physicist, Environmental Laboratory.}
and Kevin R. Pilkiewicz\footnote{Research Physicist, Environmental Laboratory.}}
\affil{Environmental Laboratory \\ U.S. Army Engineer Research and Development Center (ERDC), Vicksburg, MS 39180, USA.}

\begin{document}

\maketitle

\begin{abstract}
The escalating risk of collisions and the accumulation of space debris in Low Earth Orbit (LEO) has reached critical concern due to the ever increasing number of spacecraft. Addressing this crisis, especially in dealing with non-cooperative and unidentified space debris, is of paramount importance. This paper contributes to efforts in enabling autonomous swarms of small chaser satellites for target geometry determination and safe flight trajectory planning for proximity operations in LEO. Our research explores on-orbit use of the You Only Look Once v5 (YOLOv5) object detection model trained to detect satellite components. While this model has shown promise, its inherent lack of interpretability hinders human understanding, a critical aspect of validating algorithms for use in safety-critical missions. To analyze the decision processes, we introduce Probabilistic Explanations for Entropic Knowledge extraction (PEEK), a method that utilizes information theoretic analysis of the latent representations within the hidden layers of the model. Through both synthetic in hardware-in-the-loop experiments, PEEK illuminates the decision-making processes of the model, helping identify its strengths, limitations and biases.

\end{abstract}

\section{Nomenclature}

{\renewcommand\arraystretch{1.0}
\noindent\begin{longtable*}{@{}l @{\quad=\quad} l@{}}
AP & average precision \\
APF & artificial potential field \\
CAM & class activation map \\
CNN & convolutional neural network \\
FPN & feature pyramid network \\
GPU & graphics processing unit \\
Grad-CAM & gradient-weighted class activation map \\
$H(X)$ & entropy of a random variable $X$ \\
IoU & intersection over union \\
LEO & Low Earth Orbit \\
mAP & mean average precision \\
mAP@0.5 & mewan average precision at 0.5 IoU threshold \\
mAP@0.5:0.95 & mean average precision averaged over 0.5, 0.55, ..., 0.95 IoU thresholds \\
OOS & on-orbit servicing \\
P & precision \\
PANet & path aggregation network \\
pdf & probability density function \\
PEEK & Probabilistic Explanations for Entropic Knowledge extraction \\
PR curve & precision-recall curve \\
R & recall \\
R-CNN & region-based convolutional neural network \\
ReLU & rectified linear unit activation function \\
SWaP & size, weight, and power \\
RPN & region proposal network \\
TP & true positive \\
YOLO & You Only Look Once
\end{longtable*}}

\section{Introduction}
\lettrine{T}{he} increasing number of spacecraft in Low Earth Orbit (LEO) from governments and private companies has delivered immense benefits, including high-precision remote sensing of the Earth, microgravity and space environment research, and deep space imaging. However, this orbital crowding has precipitated growing risks of collisions. In-space missile tests carried out by multiple nations \cite{Johnson2021, Kestenbaum2007, Henry2019, SCPAO2021} have scattered space debris that remains in LEO, and many decommissioned or dysfunctional spacecraft are also still in orbit. This has become such an acute problem that the UN General Assembly passed a resolution in late 2022 calling for a halt to destructive direct-ascent anti-satellite missile testing \cite{UN2022}.

One approach to reduce the future proliferation of space debris is to extend the lifetimes of satellites with on-orbit servicing (OOS). Previous missions demonstrating on-orbit operations include DART \cite{Cheng2018}, XSS-10 \cite{davis2003xss}, XSS-11 \cite{AFRL2011}, and ANGELS \cite{AFRL2014}. More recently, Northrop Grumman performed OOS operations around IS-901 \cite{Sheetz2020} and IS-10-02 \cite{NorthropGrumman2021}. However, these missions are operated around a known RSO with a single chaser. In reality, the greatest collision risks are posed by large, non-cooperative RSOs whose structure, geometry, and capabilities are not well known. At the same time, close proximity operations are challenging: manned missions to such unknown environments are dangerous; lag times in data transfer make real-time, ground-based analysis of the target during approach infeasible; and on-board computers have limited capabilities. The stakes are high–uncontrolled collisions between the approaching spacecraft and target would exacerbate the debris problem and could cause significant damage and injury. A promising solution is to use an autonomous swarm of small chaser satellites to perform operations around the target. This would substantially lower the risks required to de-tumble and perform other OOS operations around non-cooperative spacecraft. This requires accurate and trustworthy machine vision and sensing systems that can run in low-compute environments onboard the swarm.

Computer vision has become exceptionally powerful in recent years. Even challenging missions like autonomous rendezvous and docking with non-cooperative satellites could be feasible if modern vision algorithms can be harnessed for use in space. There are two primary barriers: spaceflight hardware is much less powerful than the large conventional computers typically used for state-of-the-art computer vision and the opaque decision processes of these models make validation difficult. Algorithmic developments have decreased computational costs while low size, weight, and power (SWaP) computers have improved. Prior work has sufficiently narrowed the gap in accuracy and runtime to permit satellite component recognition to run on spaceflight-equivalent hardware. However, the problems of interpreting model decisions and, consequently, building trust in these techniques for high-stakes operations is largely unsolved.

This article aims to address this problem by developing a method to provide explanations of a lightweight satellite component detection algorithm's decision process. We leverage information theoretic analysis of the input data and the model's internal latent data representations to efficiently generate explanations of the decision process in visual and probabilistic terms. This provides insight into failure modes, enables more robust validation, and will guide future model developments.

The major contributions of this work are:

\begin{enumerate}
    \item Probabilistic Explanations for Entropic Knowledge extraction (PEEK), an efficient technique for explaining CNN decisions by cross-referencing empirical entropy estimates of internal data representations with input images.
    
    \item Hardware-in-the-loop experiments on video feeds of a real-world satellite mock-up under realistic lighting and motion conditions using a YOLOv5-based satellite component detector.
    
    \item An in-depth analysis of failure modes of effective satellite component detector algorithms and attribution of failures as true risks or artifacts of the experimental environment.
\end{enumerate}

\section{Related Literature}

This research draws significantly from existing literature, including work in computer vision such as vision systems deployed onboard spacecraft, and the mathematical field of information theory.

\subsection{Computer Vision}

Over the past 15 years, deep CNNs have revolutionized computer vision, overtaking traditional algorithms in nearly all vision tasks. Several factors converged resulting in these gains. Large, annotated image datasets intended for training were made public \cite{Deng2009, lin2014microsoft}). GPU acceleration \cite{Krizhevsky2017} enabled extreme parallelization of neural computations, which then enabled much larger, more expressive neural architectures to be trained \cite{simonyan2014very, Szegedy2015, he2016deep}. Rapid gains in image classification prompted the community to focus on more challenging computer vision tasks, such as image segmentation, object detection, and object tracking.

Object detection simultaneously identifies multiple objects in an image, classifies them, and localizes them with tight rectangular bounding boxes with horizontal bases (illustrated in the figure below).  In this case, the process has detected two solar panels, the body, and antenna; labeled them properly; and localized them with tight bounding boxes. For each object, the algorithm predicts $(x,y)$ coordinates for the center of the bounding box, the dimensions of the box, a class prediction (solar, body, antenna, or thruster in this case), and a score measuring how confident the model is that it has located an object (see the numbers with the predicted class labels in the diagram).

\begin{figure}[H]
    \centering
    \includegraphics[width=0.6\textwidth]{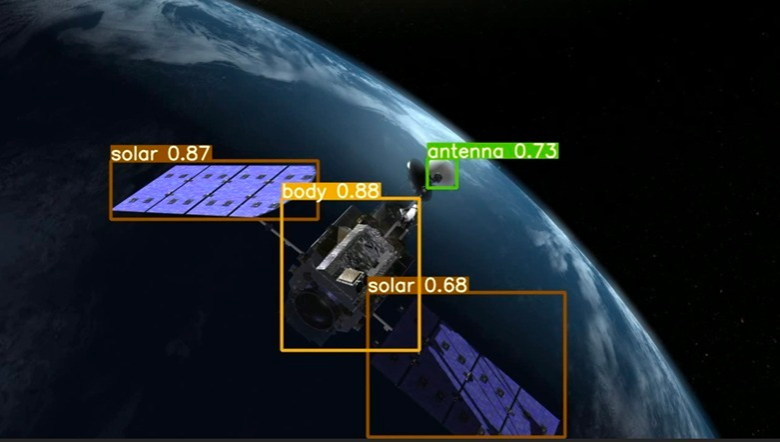}
    \caption{Object detection performed on satellite components}
    \label{fig:my_label}
\end{figure}

Early algorithms \cite{Szegedy2013deep} simply performed brute-force searches for objects–they extracted all bounding boxes along sliding windows of different sizes, fed the extracted sub-windows to high-quality CNN image classifiers, selected those with high classification confidence, and then pruned away redundant detections. This approach soon gave way to algorithms designed specifically for object detection. The multi-stage character of leading object detection algorithms was eroded over time to reduce computational costs and improve accuracy. Region-based CNNs (R-CNNs) \cite{Girshick2013} deployed selective search to generate intelligent region proposals where objects may exist, leading to far fewer, high-quality bounding boxes to be fed into CNN classifiers. OverFeat \cite{Sermanet2013} and SPPNet \cite{he2015spatial} sped up the feature extraction with new convolutional techniques and pool features in arbitrary regions. Faster R-CNN \cite{Ren2015} replaced selective search with a region proposal network (RPN), a CNN that proposes regions. The RPN was paired with the CNN classifier, and the two were trained in alternating stages, resulting in better-quality object detections with lower inference times. Single-stage detectors go a step further and perform the entire task in one network that maps raw input image pixels to bounding boxes, class probabilities, and confidence (objectness) scores simultaneously. These algorithms run an order of magnitude faster than multi-stage networks but were initially plagued by poor accuracy. The You Only Look Once (YOLO) \cite{Redmon2015, Redmon2016, Redmon2018YOLOv3, bochkovskiy2020yolov4, ge2021yolox, Jocher2022, li2022yolov6, wang2022yolov7, Jocher2023YOLOv8} family of single-stage detectors have eliminated the gap in performance in recent years, now competing with the best multi-stage object detectors in a single-stage architecture.

We experimented with lightweight versions of several variations of YOLO to address the gap in the literature regarding performance comparisons. YOLOv5s, YOLOv6n, YOLOv7t, YOLOv8n, and YOLOv8s were tested in our experiments. YOLOv5s and YOLOv8s yielded the best accuracy on a test split of synthetic image data. However, YOLOv5s generalized significantly better to satellites unseen during training than YOLOv8s in hardware-in-the-loop experiments.

\subsection{On-board Computer Vision for In-Space Applications}

Prior work by the authors have explored computer vision and sensor fusion for onboard spacecraft component detection, 3D reconstruction of spacecraft \cite{caruso2023} from inspection orbits, and guidance and navigation for close proximity operations. This research demonstrated YOLOv5 \cite{Jocher2022} can effectively detect satellite bodies, solar arrays, antennas, and thrusters \cite{mahendrakar2021real, mahendrakar2023impact} on a low compute budget (Raspberry Pi with Intel Neural Compute Stick 2), both in testing images of arbitrary spacecraft and in a lab setting with a satellite mock-up not seen during training. Multi-stage detector Faster R-CNN displayed high object detection accuracy but could not run at frame rates sufficient for GNC applications. \cite{mahendrakar2022performance}. Video feeds were later fused with depth data estimated by a stereographic camera to build 3D point-clouds marking bodies, solar arrays, and antennas. Artificial potential field (APF) guidance algorithms marked bodies as attractive and fragile components as repulsive. It successfully guided software-simulated \cite{mahendrakar2021use} chaser spacecraft around fragile solar panels and antennas to dock safely with bodies of non-cooperative spacecraft. Later flight tests with drones \cite{mahendrakar2023autonomous} demonstrate this approach to autonomous rendezvous with non-cooperative spacecraft is feasible, though ongoing research aims to further improve model accuracy and run times.

However, CNN decision-making remains complex and difficult for humans to grasp fully. With such high human and financial costs involved with failure, a model providing no explanations for its decisions is not likely to be adopted for these safety-critical on-orbit missions. In addition, it is difficult to develop remedies and introduce safeguards into the algorithm without a better understanding of why failures occur. These concerns call for further study into how, when, and why models fail.

\subsection{Visualization of Neural Network Predictions}

One of the most commonly used methods for CNN classifier visualization is gradient-weighted class activation mapping (Grad-CAM) \cite{selvaraju2017grad}, which measures and visualizes the importance of each pixel of the input image to the CNN's classification decision. For a given target class and an image, it performs a forward pass with the CNN and extracts the class activation maps (CAMs) of the final convolutional layer. This layer keys in on high-level features used in the CNN's decision process. Grad-CAM then finds the gradient of the target class score with respect to CAMs from earlier layers using a backward pass. These gradients are pooled spatially across the CAMs. After applying the ReLU activation and upscaling using guided backpropagation \cite{Springenberg2015}, the result is overlaid as a heatmap on the input image, where the higher intensities correspond are pixels that were most influential to the class prediction.

Grad-CAM and several other visualization methods (e.g., Grad-CAM++ \cite{8354201}, CNN fixations \cite{mopuri2018cnn}, DeepLIFT \cite{shrikumar2017learning}) all fall under the class discriminative methods. The performance of these methods heavily rely on the classification being accurate. If objects are classified incorrectly, the visualizations will be distorted or incorrect, defeating the purpose of visualizing the CNNs decision process. To overcome the challenges of the class discriminative methods, Eigen-CAM \cite{muhammad2020eigen} works based entirely on the forward pass. It assumes that during the training process of the CNN, all non-relevant features learned from the training data will be regularized or smoothed out and only relevant features are preserved in the model, hence eliminating the dependency on classifier performance and providing class-agnostic analysis of CNN decisions.

Eigen-CAM uses singular value decomposition (SVD) to compute the principal components of the feature maps and projects the input image onto the last convolutional layer. The class activation map is then generated by projecting this output onto the first eigenvector. While Eigen-CAM does not need to compute a expensive backward pass, the SVD can still be computationally intensive.

\subsection{Entropy and Information Theory}

Information theory is the study of information within data signals \cite{Cover2006}. Of primary interest to this work are input signals, i.e. images or video feeds, and signals passing through the hidden layers of neural networks. Entropy is an information theoretic measure of uncertainty in the distribution of a random variable $X$. The (differentiable) entropy of a continuous random variable $X$ valued in $\mathbb{R}^k$ with probability density function (pdf) $f$ is defined as
\begin{align}
    H(X)=\mathbb{E}\left[-\log f(X)\right]=-\int_{\mathbb{R}^k} f(x)\log f(x)\,dx.
\end{align}

Recent work \cite{meni2023entropy} established the importance of entropy propagation through CNNs to their performance. Hence, we hypothesize the entropy of the internal feature maps of a CNN computed during a forward pass can help explain the model's decisions.

\section{Motivation and Methods}

\subsection{Patterns in CNN Latent Representations}

\begin{figure}[H]
    \centering
    \includegraphics[width=\textwidth]{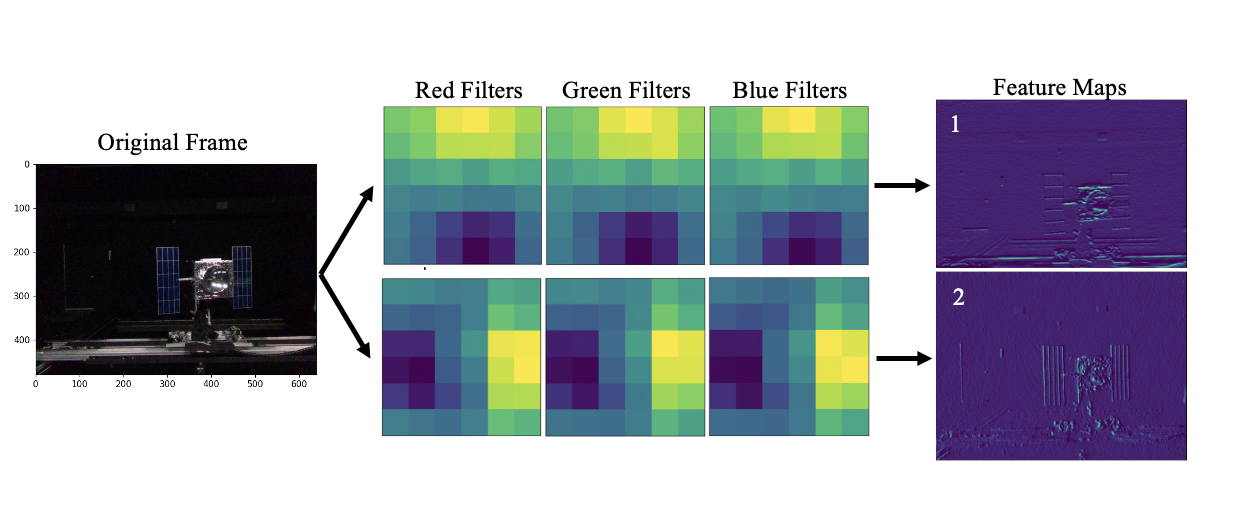
    }
    \caption{Features extracted by convolutional filters of the first layer of YOLOv5}
    \label{feature map visualition}
\end{figure}

Inside CNN layers are convolutional filters that, when applied to an input image, compute a feature map. More specifically, the weights inside the filters are learned during model training to extract useful features such as edges and color to create these feature maps, which are useful for the downstream task of object detection. These features are stored as hidden representations of the input data, called feature maps. An example of this process on can be seen in the figure below, where we extracted two $6\times 6\times 3$ convolutional filters (the two rows of in Figure~\ref{feature map visualition}: one $6\times 6$ slice applied to each color channel) and the feature maps they produce from the first convolutional module of YOLOv5. Figure~\ref{feature map visualition} shows the impact of the filters to achieve specific feature maps.

YOLOv5 was trained to perform object detection by minimizing classification and localization errors. During training, these two filters learned to become edge detectors despite no explicit guidance by the optimization process. In particular, the first filter in Figure~\ref{feature map visualition} has negative weights (darker) on the bottom of the filter and positive weights (lighter) on the top, both with more intensity near the center. This combination of blurring (more muted values in the corners) and horizontal dividing line between positives and negatives is equivalent to the classical top Sobel filter \cite{marr1980theory} for detecting horizontal lines. Note the horizontal lines on the solar panels are easily visible but the vertical lines are not. The resulting Feature Map 1 clearly enhances horizontal lines. Similarly, the second filter is similar to the classical right Sobel filter for detecting vertical lines, and we see the vertical lines on the solar panel on Feature Map 2 in Figure~\ref{feature map visualition}.

The collection of feature maps in a convolutional layer contain the latent representations of the data as it flows through the network (as seen above). This is where we will be performing information theoretic analysis because they contain signals with information content that helps us to perform the downstream task.

\subsection{PEEK}

A pixel-level entropy measure applied to the feature maps will be used to analyze the latent data representations within the YOLOv5 architecture and its internal feature maps. This analysis adds transparency to the model's decision-making process and provides an avenue to detect failure modes and offer remedies for them.

Our method, Probabilistic Explanations for Entropic Knowledge extraction (PEEK), computes a depth-wise entropy-related measure at each ``pixel'' of the latent feature maps of hidden convolutional layers of CNNs. These latent representations $L\in\mathbb{R}^{l\times w\times d}$ are stacks of feature maps of shape $l\times w$ extracted by the convolutional filters in the layer. We first shift the ``pixel'' intensities of each feature map to be positive:
\begin{align}
L^\text{pos}_{ijk}=L_{ijk}+\left|\min\limits_{\{i\leq l, j\leq w\}}L_{ijk}\right| \label{positivizing}
\end{align}
Since all feature maps tend to have both positive and negative values, formula \eqref{positivizing} shifts the values to the positive scale for each feature map separately. This both preserves the relative differences between pixels within a feature map and retains the strength of intense feature maps. PEEK then computes the entropy ($g(x)=-x\ln x$ as defined in the convex optimization context \cite{boyd_vandenberghe_2004}) of each depth-wise column of pixels in spatial location $(i,j)$, resulting in the PEEK intensity at that location:
\begin{align}
P_{ij}&=-\sum\limits_{k=1}^d g\left(L^{\text{pos}}_{ijk}\right)
\end{align}
We apply this PEEK calculation to each spatial location in the hidden representation $L$ to compute a full PEEK map for the selected convolutional layers of the neural network. This PEEK map can be resized and overlaid on the input image, resulting in highlighting of high-entropy locations in the feature maps corresponding to that location.

In this paper, we will employ PEEK and Eigen-CAM to visualize the latent data representations within a YOLOv5s model trained for satellite component detection. This will enable a direct comparison of visualization techniques and their effectiveness.

\section{Performance Evaluation and Experimental Conditions}

This section specifies the object detection performance metrics, the datasets used for training and testing the algorithms, the lab setting for hardware-in-the-loop experiments.

\subsection{Object Detection Metrics}

The model's performance is measured in terms of its ability to classify and localize features. This paper evaluates and compares the mean average precision (mAP) at the intersection over union (IoU) thresholds 0.5 and 0.5:0.95. IoU measures the overlap between two bounding boxes by dividing the area of their intersection by the area of their union. It ranges from 0 (no overlap) to 1 (full overlap).

\begin{align}
    IoU &= \frac{\text{Area of intersection of bounding boxes}}{\text{Area of union of bounding boxes}}
    \end{align}
A predicted class and bounding box is called a true positive (TP) in object detection if the bounding box has IoU with a ground-truth bounding box above a specified IoU threshold and it is classified correctly. Precision (P) is the ratio of TP detections to all positive detections made by the model, which descries how reliable model predictions are. Recall (R) is the ratio of TP detections to all positive instances in the dataset, which describes how well the model detects objects in images.

\begin{align}
    P &= \frac{TP}{\text{Positive object predictions}}\\
    \notag \\
    R &= \frac{TP}{\text{Ground-truth object instances}}
\end{align}

A precision-recall (PR) curve is generated by plotting precision versus recall for a range of confidence thresholds required to make detections for each class. Average precision (AP) for a specified class is the area under the PR curve for that class. Mean average precision (mAP) is the mean of class-specific APs. The resulting mAP is a one-number summary of object detection quality for a specified IoU threshold.

\subsection{Training Dataset}
\begin{figure}[H]
    \centering
    \includegraphics[width=0.7\textwidth]{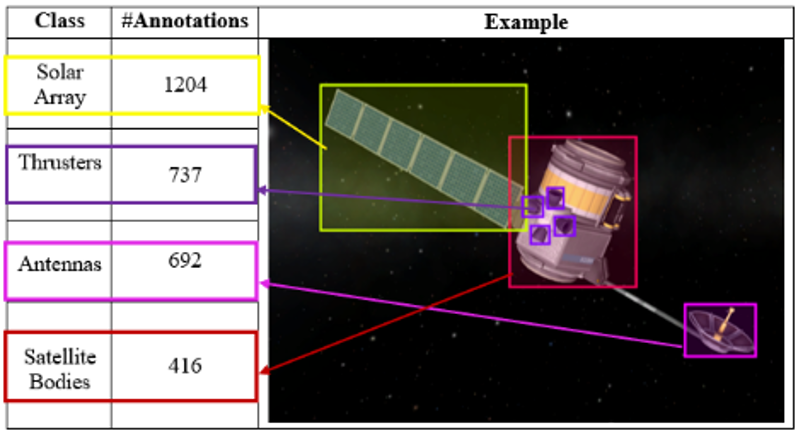}
    \caption{Training Dataset}
    \label{training_dataset}
\end{figure}

The training data consists of images of spacecraft from Google, mostly generated by STK and Kerbal Space Program as well as Blender-generated images of NASAs stereo lithography models. A total of 1231 images are collected for the training dataset. Figure~\ref{training_dataset} shows the total number of annotations present in the dataset for each component.

\begin{figure}[H]
    \centering
    \includegraphics[width=0.7\textwidth]{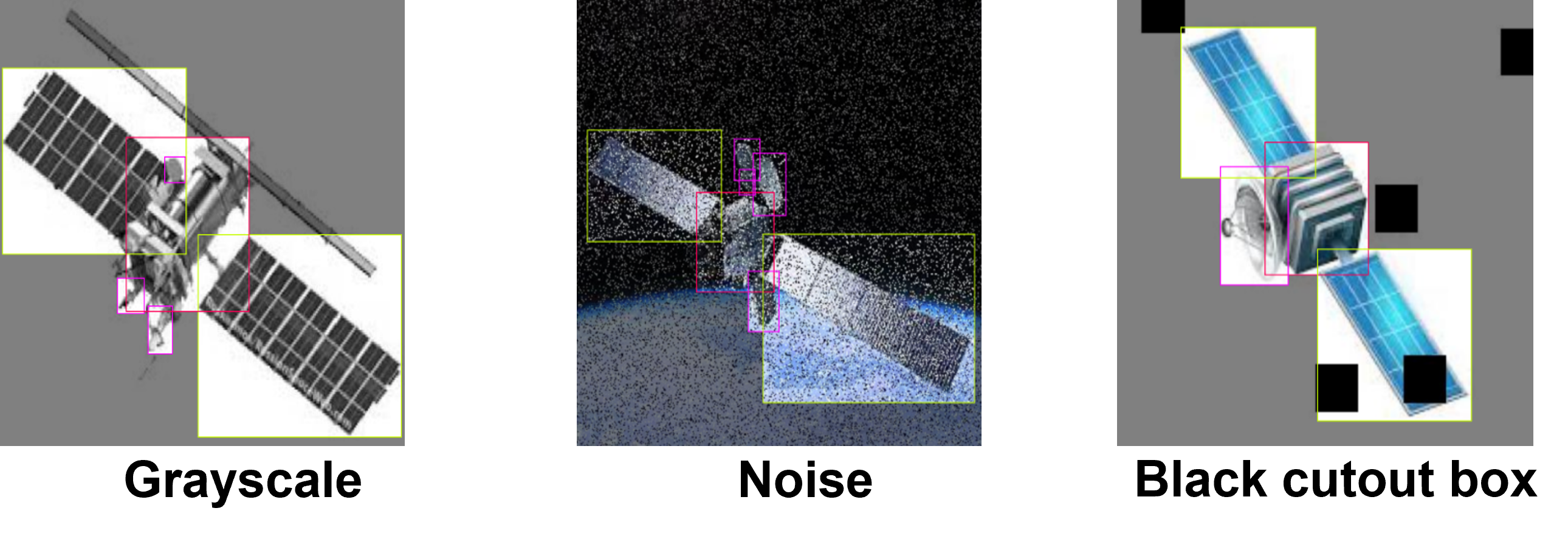}
    \caption{Training Dataset Augmentation}
    \label{augmentation}
\end{figure}

This dataset was randomly split into a train, validation, and test subsets (60\%/20\%/20\%). YOLOv5s was trained on the training split using image augmentation to include rotation, black cut-out boxes, Gaussian noise, and greyscale based on camera issues. These are shown in Figure~\ref{augmentation}.

The validation subset was used for tuning the model. The tuned model achieves 0.587 mAP@0.5 on the test subset not seen during training. More rigorous testing was then performed in the lab as discussed next.

\subsection{Testing Dataset and Experimental Setting}

\begin{figure}[H]
\centering
\begin{subfigure}{.5\textwidth}
  \centering
  \includegraphics[width=.9\linewidth]{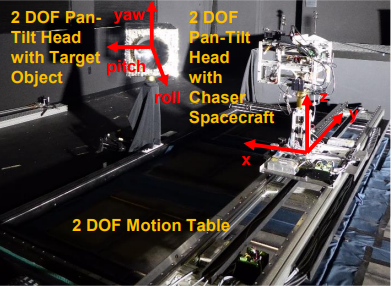}
  \caption{Target satellite mock-up and chaser simulator}
  \label{target_simulator}
\end{subfigure}%
\begin{subfigure}{.5\textwidth}
  \centering
  \includegraphics[width=.9\linewidth]{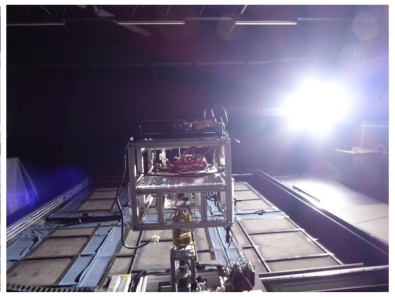}
  \caption{Testbed simulation of extreme lighting conditions}
  \label{lighting_conditions}
\end{subfigure}
\caption{The ORION Testbed at the Autonomy Lab at Florida Tech}\label{ORION}
\end{figure}

This paper uses an experimental testing dataset \cite{attzs2023comparison} that includes hardware-in-the-loop image frames extracted at 30 frames per second from two videos of a mock-up satellite shot under extreme lighting conditions. The mock-up satellite yaws at a constant angular rate of 1.5 deg/second for 3 minutes on the ORION test bed \cite{wilde2016orion}  at the Autonomy Lab in Florida Tech. The ORION testbed depicted in Figure~\ref{ORION} enables us to create in-space visual conditions suitable for testing this algorithm. The test bed is equipped with a planar maneuver kinematics simulator that accommodates pan-tilt mechanisms custom-built to simulate on-orbit proximity operations used to simulate chaser satellite motion, as seen in Figure~\ref{target_simulator}. The simulator uses a separate stationary pan-tilt head to simulate the attitude motion of a target satellite model. The model features geometry and surfaces typically found on satellites, such as solar panels, parabolic antennas, body panels, and thruster nozzles. All the walls, floors, ceiling and windows panels of the room are covered with high absorptive black paint and black curtains. A Hilio D12 LED LitePanel capable of generating light with a color of 5,600K (daylight balanced temperature) is used to simulate space-like lighting conditions, as seen in Figure~\ref{lighting_conditions}. 

\begin{figure}[H]
    \centering
    \includegraphics[width=0.95\textwidth]{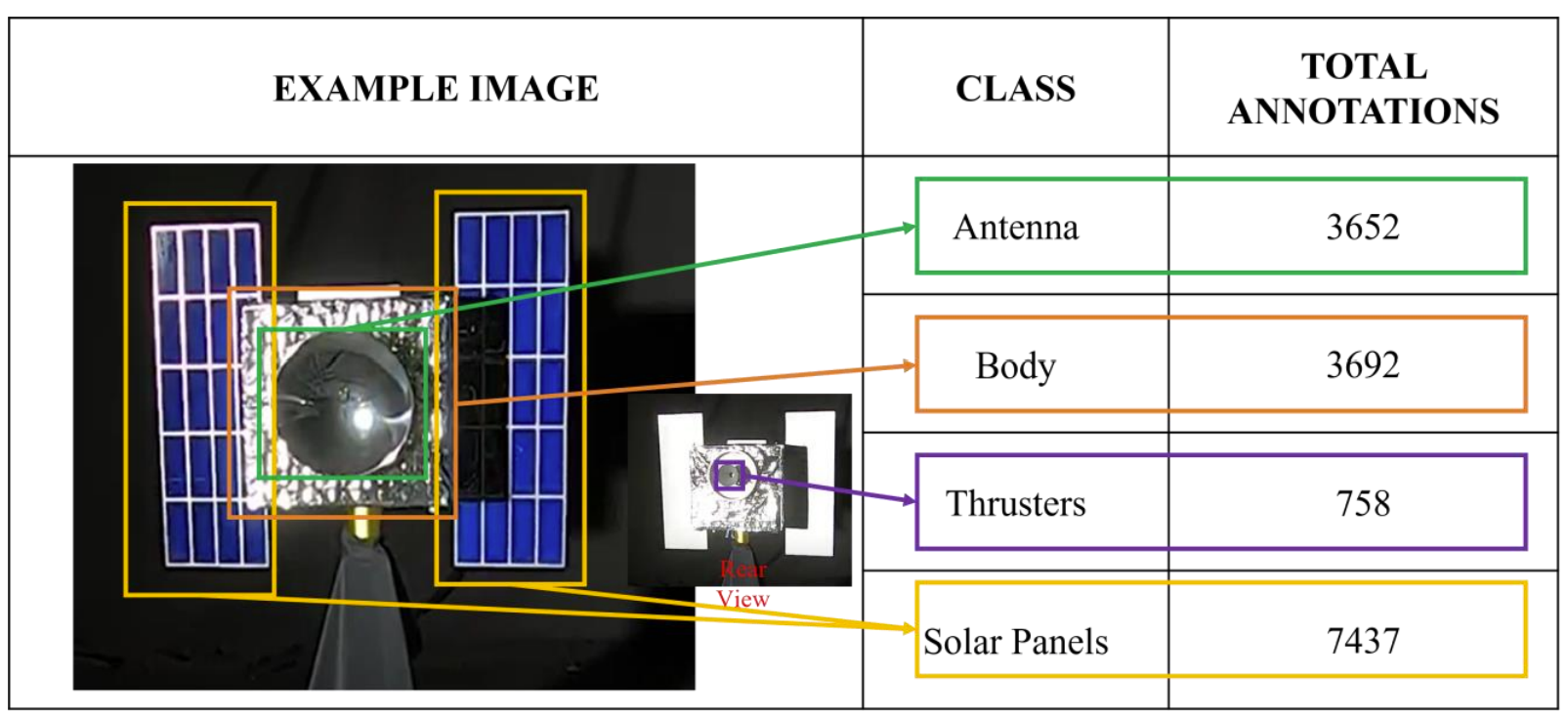}
    \caption{Satellite mock-up and its features}
    \label{satellite_mockup}
\end{figure}

Figure~\ref{satellite_mockup} depicts the satellite mock-up with its features. It contains 1800 frames annotated with its two solar panels, thruster, body, and antenna. The object detectors used below attempt to detect each visible feature in each frame of the video.

The trained YOLOv5s model achieves 0.377 mAP@0.5. The performance is degraded in this real-world setting, but this remains relatively good performance when compared to the object detection literature.

\section{Experimental Results}

\subsection{Architecture Analysis via PEEK}

\begin{figure}[H]
    \centering
    \includegraphics[width=\textwidth]{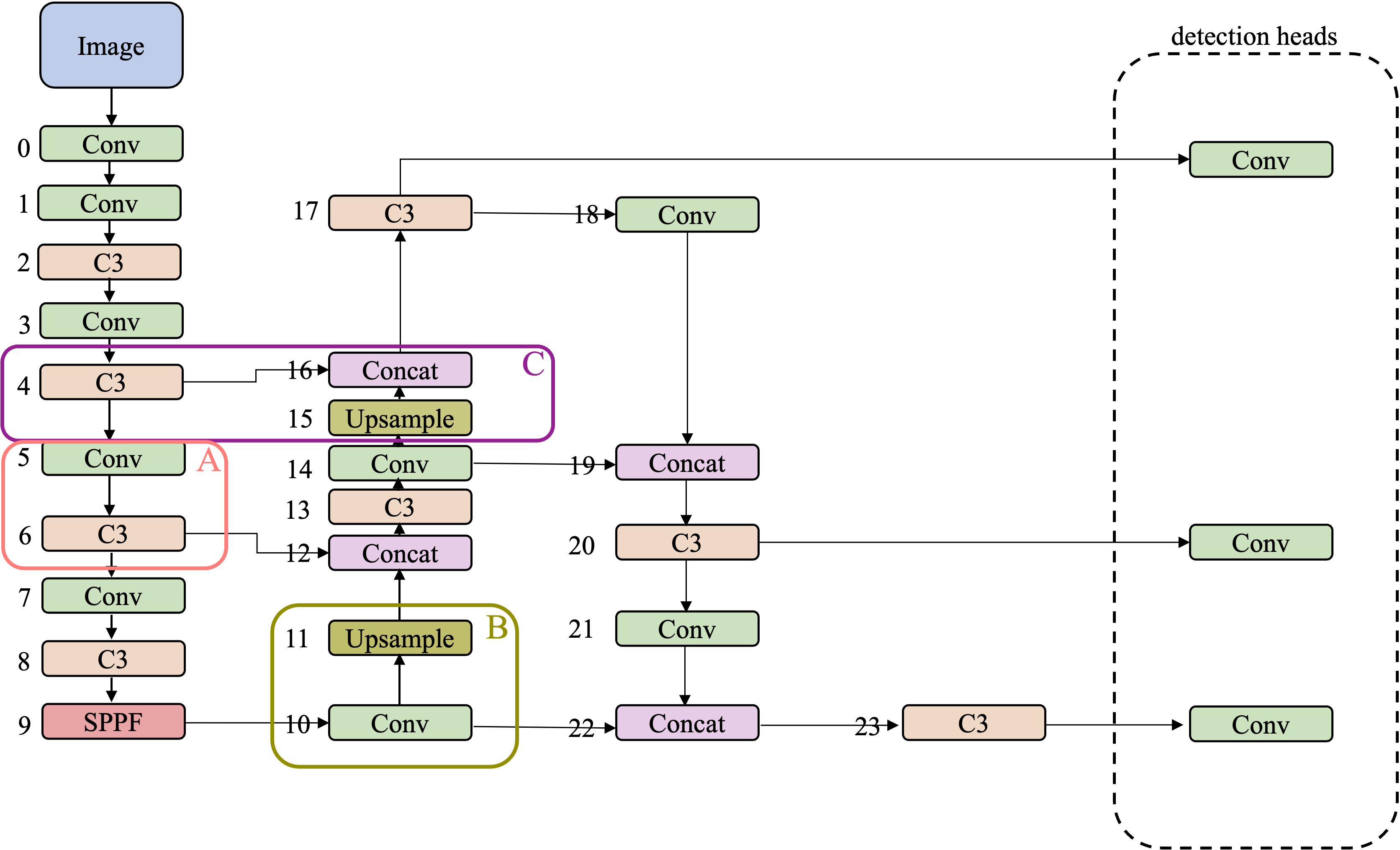}
    \caption{YOLOv5 (v6.0/6.1) Architecture}
    \label{fig:yolov5}
\end{figure}

The architecture of YOLOv5 as shown in Figure~\ref{fig:yolov5} is made up of 24 modules separated into the backbone, neck, and head. The backbone (modules 0-9) consists of strided convolutional layers and C3 modules, adapted from \cite{wang2020cspnet} and optimized).  As the convolutional filter scans over the input, it extracts patterns from image patches spanning the entire input but represents them in a lower dimension in the next layer. This shrinks the feature map shapes by half while still representing the entire image, so each pixel of the output represents a larger image patch. The next layer scans over those output feature maps so they extract patterns from regions that represent more space in the original image. Therefore, this portion of the backbone extracts features at successively larger spatial scales. Last in the backbone is spatial pyramidal pooling ``fast'' (SPPF) \cite{he2015spatial}, which pooling at different scales, helping the model capture features useful for detecting objects of various sizes.

The neck (modules 10-23) synthesizes features extracted at different scales and with different levels of abstraction. Modules 10-17 work with the backbone to form a feature pyramid network (FPN) \cite{lin2017feature}. The FPN treats the successively smaller feature maps learned by the backbone as the levels of the pyramid. The bottom of the pyramid corresponds to larger feature maps that have high enough resolution to encode small, fine-grained features. As we move up the pyramid, the model starts to learn larger features. While the FPN focuses on features of the objects themselves, modules 18-23 form a path aggregation network (PANet) \cite{liu2018path} that helps estimate object positioning. It also shortens the path from detection heads that focus on detecting small, medium, and large-scale objects to appropriately-sized features of those objects learned by the FPN.

The head (module 24) predicts bounding boxes and class probabilities for each object identified in the input image. After performing non-max suppression to remove redundant detections, the entire architecture is trained to minimize errors for objects detected by the heads in terms of both object classification and localization by bounding boxes.

\begin{figure}[H]
\centering
    \begin{subfigure}{.25\textwidth}
      \centering
      \includegraphics[width=\textwidth]{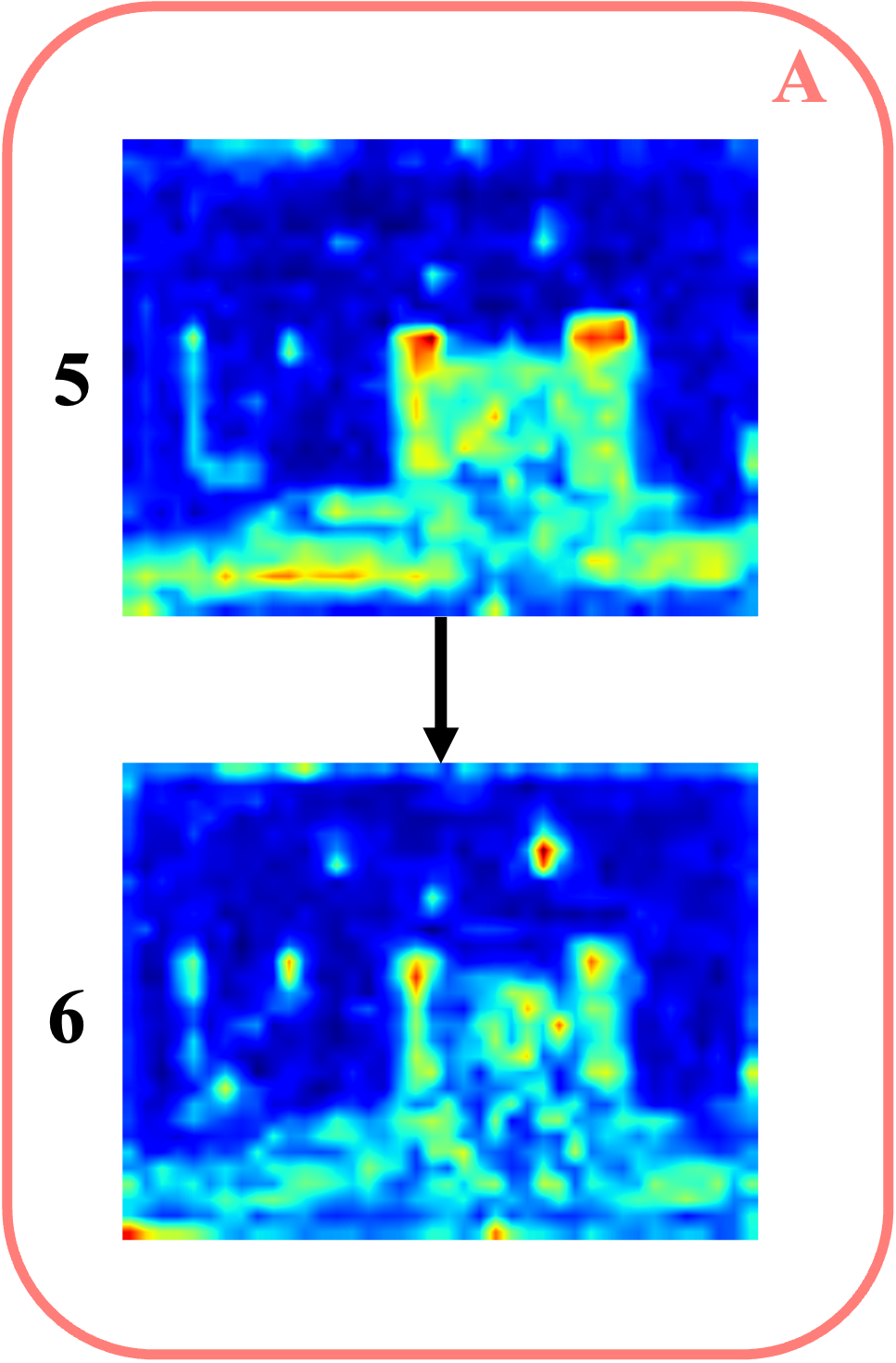}
      \caption{C3 Module}
      \label{fig:C3module}
    \end{subfigure}%
    \begin{subfigure}{.25\textwidth}
      \centering
      \includegraphics[width=\textwidth]{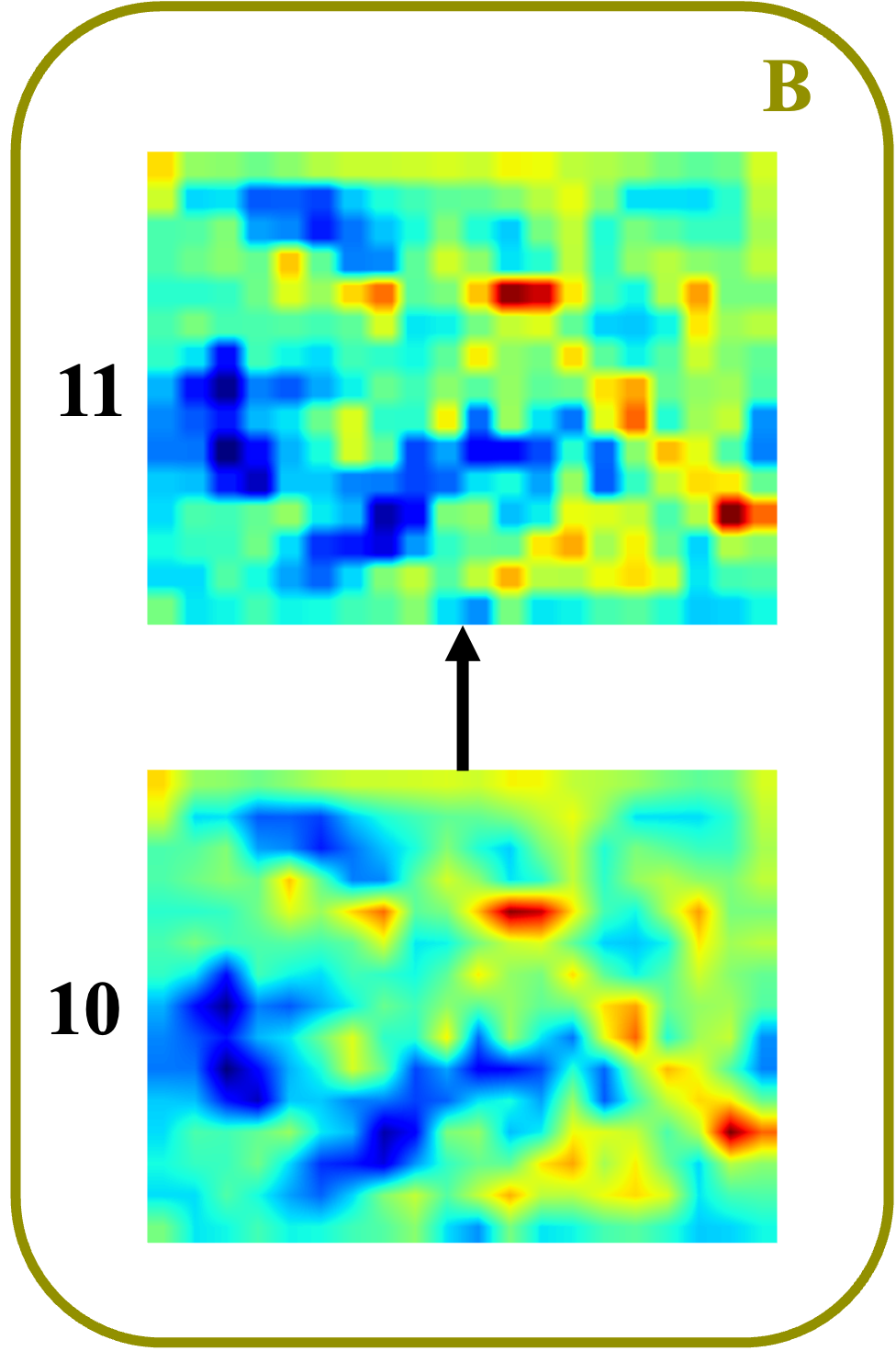}
      \caption{Upsampling}
      \label{fig:upsamplingmodule}
    \end{subfigure}
    \begin{subfigure}{.48\textwidth}
      \centering
      \includegraphics[width=\textwidth]{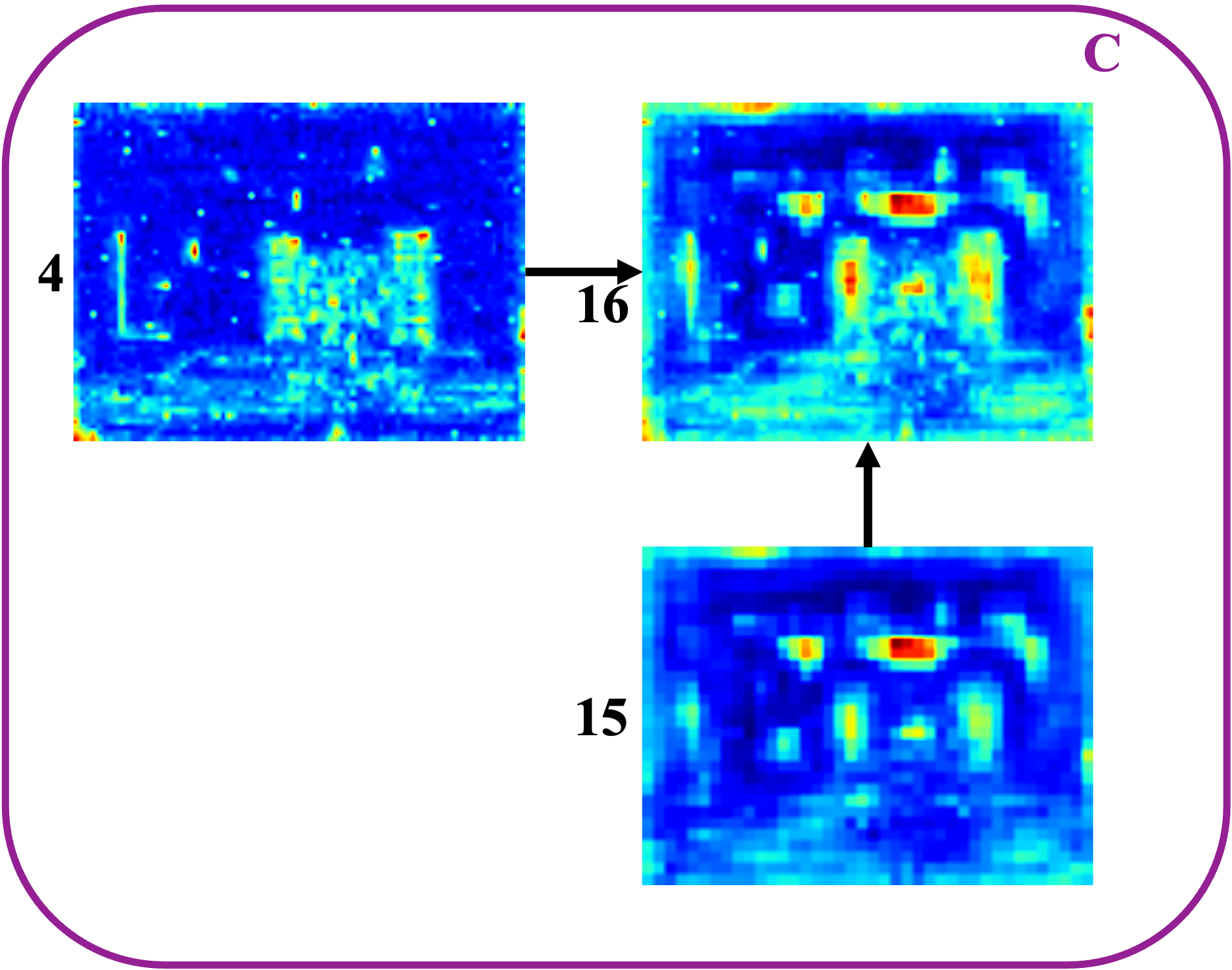}
      \caption{Concatenation}
      \label{fig:concatenation}
    \end{subfigure}
\caption{The effects of architectural structures on latent representations as revealed by PEEK}
\label{fig:test}
\end{figure}

PEEK reveals numerous intuitive results regarding the architecture of YOLOv5. In Figure~\ref{fig:C3module}, the high-entropy features in Module 5's feature maps include the satellite and its features as intended, but also the metal tracks along the ground and the door frame on the wall to the left. In addition, the satellite's solar panels, body, and antenna fade into one another, which would make disentangling the objects difficult. The C3 Module and its bottlenecks reduce the relative entropy of the extraneous features and provide clear separation between the satellite features. Upsampling Module 11 from the neck simply performs nearest neighbor upscaling to increase the resolution of the inputs. As expected, it has no impact on the PEEK features aside from making them higher resolution like the feature maps: as we see in Figure~\ref{fig:upsamplingmodule}, they just become more pixelated. Lastly, PEEK demonstrates that Concatenation Module 16 successfully fuses the well-defined features in Module 4 and more abstract features from Module 10 into intense but disentangled satellite features and well-segmented extraneous features on the back wall of the lab, as seen in Figure~\ref{fig:concatenation}.

\subsection{PEEK vs Eigen-CAM}

\begin{figure}[H]
    \centering
    \begin{subfigure}{\textwidth}
    \centering
        \includegraphics[width=0.7\textwidth]{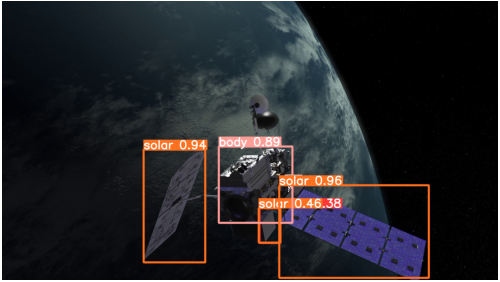}
        \caption{YOLOv5 Detections in Synthetic Video Frame}
        \label{fig:GSFC_detection}
    \end{subfigure}\\
    \begin{subfigure}{\textwidth}
        \centering
        \includegraphics[width=\textwidth]{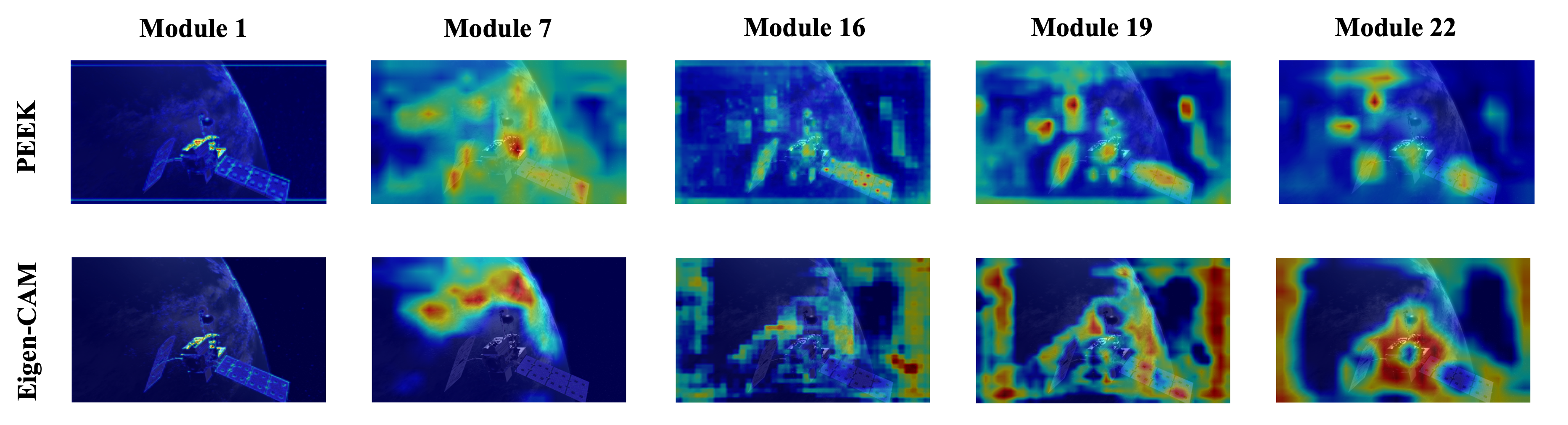}
        \caption{PEEK vs. Eigen-CAM}
        \label{fig:GSFC_comparison}
    \end{subfigure}
    \caption{PEEK Analysis of YOLOv5 Satellite Component Detector on a Synthetic Image}
\end{figure}

PEEK and Eigen-CAM both extract hidden representations computed in a forward pass through the network and provide visual cues into the model's detection process. In Figure \ref{fig:GSFC_detection} is a frame from a synthetic satellite video with YOLOv5's detections annotated. In order to analyze how the algorithm made its decisions, we compare Eigen-CAM and PEEK maps of 5 modules from the model's architecture in Figure \ref{fig:GSFC_comparison}. Both methods were prone to distraction by the Earth in the image, however, PEEK was far less impacted. 

In Module 1, both Eigen-CAM and PEEK do a good job of finding edges, specifically in the satellite itself. Yet in Module 7, both are being fooled by the earth. However, PEEK's strongest values in red are seen to be only on the satellite while the opposite is true for Eigen-CAM. The further we move through the algorithm, PEEK starts to clearly show the algorithm's focus that led to a detection. In Modules 16 and 19, PEEK shows some distraction in space and on the Earth, but is most strongly focused on the satellite, while Eigen-CAM is lost in the boundaries of the image and has minimal focus on the satellite. Similarly, in Module 22, PEEK is focused specifically on the location of the detections, while Eigen-CAM still has focus on the areas surrounding.

These results show PEEK provides visual explanations with greater explanatory value in the task of object detection.

\subsection{PEEK Practical Uses}

PEEK has allowed us to fully visualize and understand the structure of YOLOv5, explain detections, and provide us a better visualization of the internal representations of the network.  In addition to the human interpretable ability, PEEK has many practical uses.  

\begin{figure}[H]
    \centering
    \begin{subfigure}{.5\textwidth}
      \centering
      \includegraphics[width=0.95\textwidth]{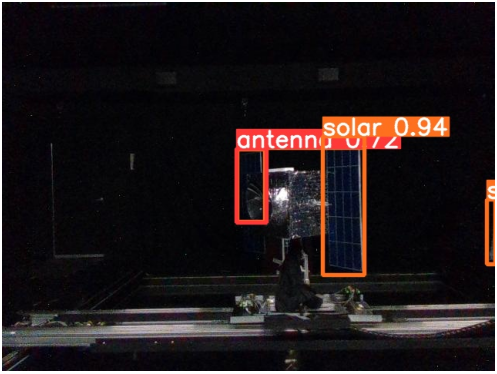}
      \caption{YOLOv5 Detections in the Lab}
      \label{fig:detections}
    \end{subfigure}%
    \begin{subfigure}{.5\textwidth}
      \centering
      \includegraphics[width=0.95\textwidth]{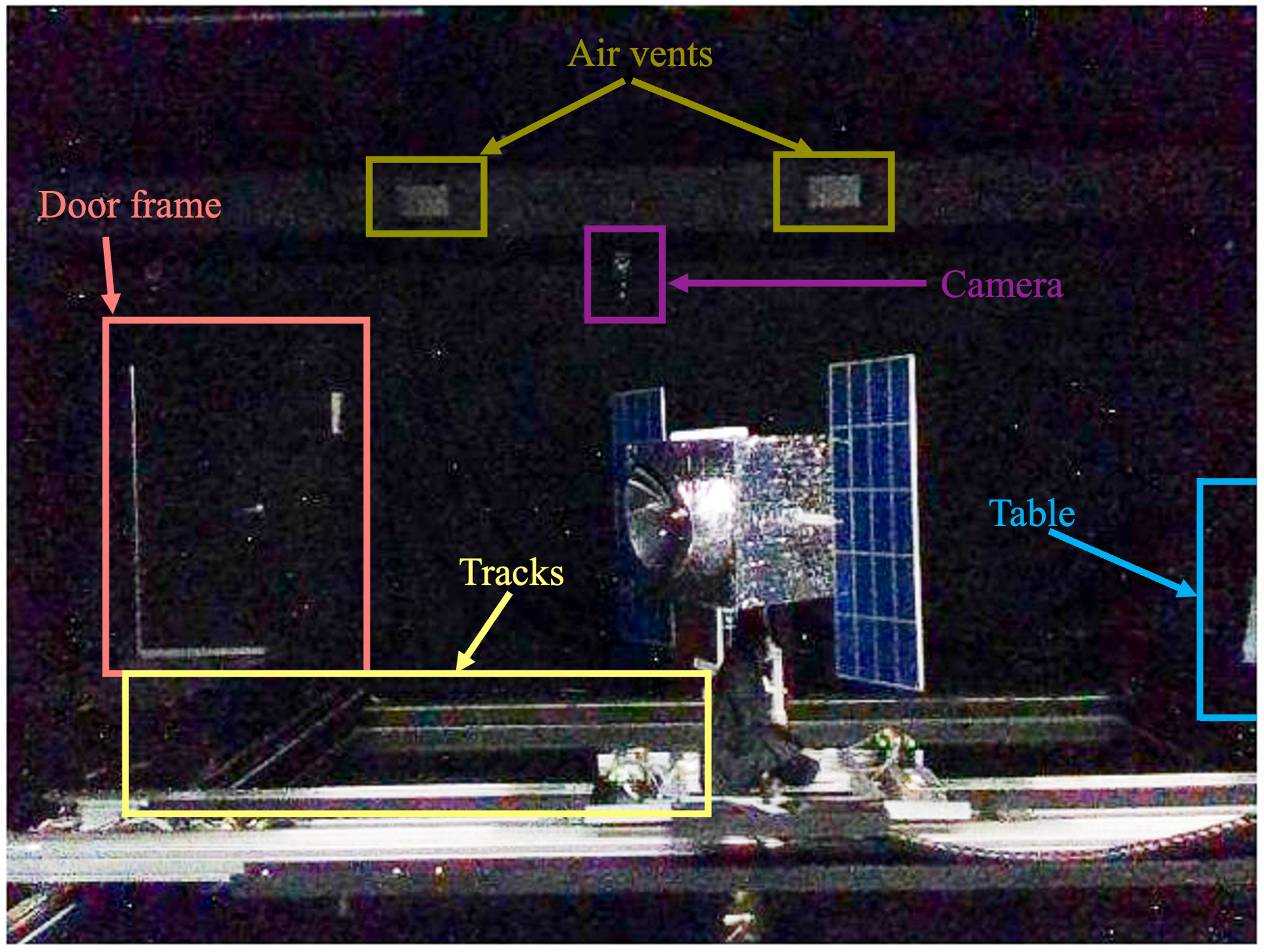}
      \caption{Brightened Image with Visible Extraneous Lab Objects}
      \label{fig:test2}
    \end{subfigure}%
    \\
    \begin{subfigure}{\textwidth}
        \includegraphics[width=\textwidth]{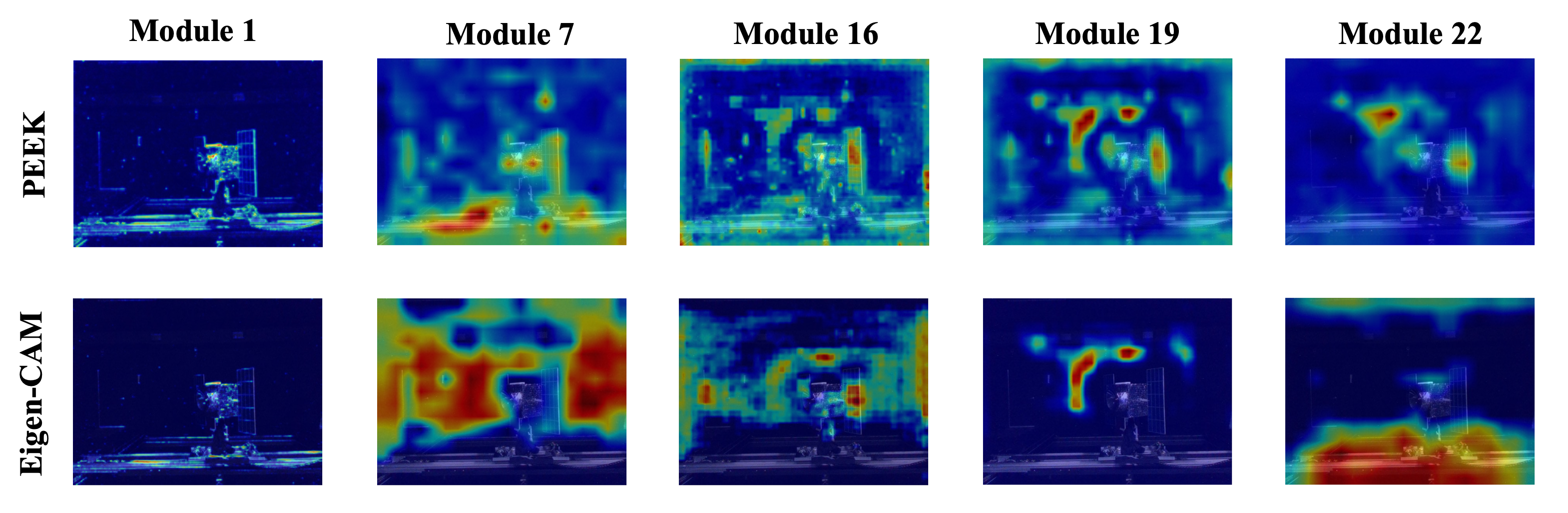}
        \caption{PEEK vs. Eigen-CAM}
        \label{fig:lab-detections}
    \end{subfigure}
\caption{PEEK for Analysis of YOLOv5 Satellite Component Detector on a Real-World Satellite Mock-Up}
\end{figure}

In Figure~\ref{fig:lab-detections}, we compare PEEK to Eigen-CAM applied to the same detection frame from our lab testing dataset. Eigen-CAM only provides minimal explanation of the detection and errors. For example, in Module 16, Eigen-CAM focuses on the satellite, along with the edges of the frame. This is where we have an incorrect detection of a solar panel on the far right of Figure \ref{fig:detections}. However, in every other Module, it provides us little to no explanation or interpretation. 

PEEK provides us explanations to possible reasons for errors in the detection along with possible issues with the data itself. In Modules 7, 16, and 19, we have high entropy visuals seen on the satellite, around the satellite in clear circles or squares, and a highlighted portion on the far right of the image. The highlighted portion on the satellite was where the algorithm correctly detected and classified a solar panel and antenna. Though the algorithm misses the body of the satellite, in PEEK's modules 19 and 22, we see that the entropy values are weak on areas around the antenna on the body, showing that it is in fact only focusing on the antenna. In Modules 16 and 19, PEEK clearly highlights a shape on the far right that is then detected and classified as a solar panel incorrectly. However, due to the highlighted shapes surrounding the the satellite, it was clear that the algorithm was picking up artifacts in the frames that were not necessarily visible to the human eye without further inspection.

Due to these artifacts seen in PEEK, we decided to take the same frame and brighten the image. Doing this revealed some aspects of our mock set up that could be confusing the algorithm. While the room is prepped to minimize the amount of light as much as possible, there are slight exposures that were only noticeable due to PEEK. In Figure \ref{fig:test2}, we identified the artifacts seen in PEEK in Figure \ref{fig:lab-detections} as air vents, a camera, a doorframe, and then a table in the far right corner which later was incorrectly detected and classified as a solar panel. This finding shows the value of PEEK and the ability to find artifacts in data, explain errors, and interpret the detections as a whole. This allows us to reevaluate our data and make changes that could improve the quality of our training, minimize biases, and prevent confusion of the algorithm.

\begin{figure}[H]
    \centering
    \includegraphics[width=0.8\textwidth]{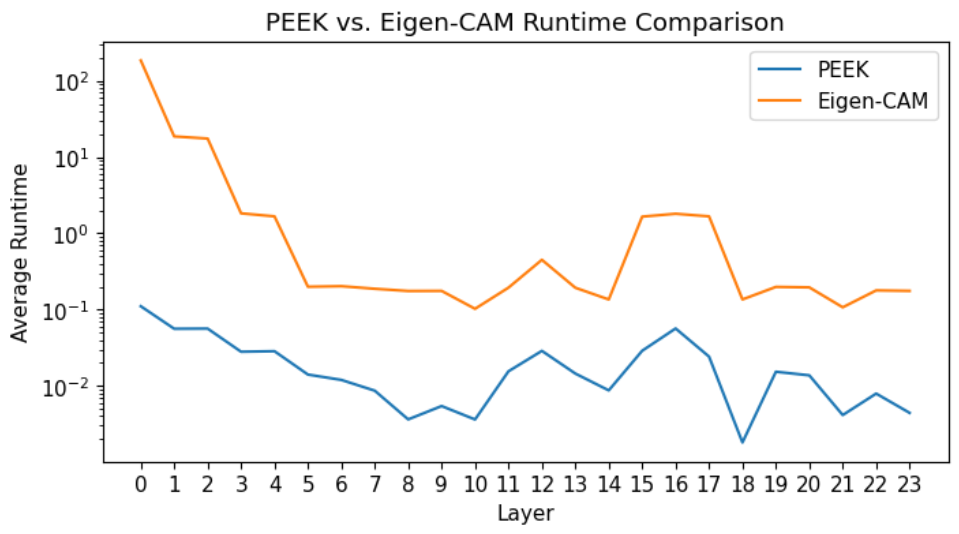}
    \caption{PEEK vs. Eigen-CAM Runtimes}
    \label{PEEK_runtimes}
\end{figure}

Lastly, the efficient PEEK calculations run far faster than Eigen-CAM. To compare them, we ran both methods 10 times, computing heatmaps for all 23 layers of YOLOv5 for the same input frames, and averaged the runtimes for each layer in Figure~\ref{PEEK_runtimes}. The compute costs rise and fall with the dimensions of the latent representations of each layer: higher at the top of the backbone in layers and top of the FPN layer 16, and lower elsewhere. Overall, PEEK runtimes are 100 to 1000 times faster than Eigen-CAM while exhibiting the benefits demonstrated above.

\section{Conclusion}

In conclusion, the growing congestion of LEO coupled with the impending risks of space debris present a pressing challenge for space agencies and private entities alike. This challenge has been exacerbated by the lack of complete awareness of all orbiting spacecraft, especially non-cooperative ones, making close-proximity operations a high-stakes endeavor. Addressing these issues requires innovative solutions, and one promising approach is the use of autonomous swarms of small chaser satellites.

However, the success of this approach hinges on the deployment of accurate and trustworthy machine vision and sensing systems that can operate in low-compute environments onboard the swarm. While computer vision has made significant strides, the unique constraints of spaceflight hardware and the opaqueness of decision-making in modern vision algorithms have presented substantial hurdles.

This article  tackled these challenges by introducing the Probabilistic Explanations for Entropic Knowledge extraction (PEEK) method, a novel technique designed to shed light on the decision process of lightweight satellite component detection algorithms. By leveraging information theoretic analysis and internal data representations, PEEK provides valuable insights into the workings of these algorithms, offering explanations in both visual and probabilistic terms. This not only aids in understanding failure modes but also facilitates robust validation and guides the future development of these critical models.

The ability to reevaluate our data in this manner offers a transformative opportunity to enhance the quality of our training datasets and experimental choices. By pinpointing sources of error and bias, we can take proactive steps to rectify these issues, ultimately leading to more accurate and reliable satellite component detection algorithms. Moreover, PEEK empowers us to prevent potential confusion in the algorithm's decision-making process, thereby ensuring the safe and efficient operation of autonomous satellite swarms in LEO.

PEEK represents a valuable asset in the ongoing quest to improve the robustness and effectiveness of machine vision systems in space and beyond.

\section*{Acknowledgments}

R. T. White thanks the NVIDIA Applied Research Accelerator Program for hardware support. The authors also thank Energy Management Aerospace, LLC, and the Oak Ridge Institute for Science and Education (ORISE) for financial support of M.J. Meni (MIPRs W81EWF30204272 and W81EWF31876813). Opinions, interpretations, conclusions, and recommendations are those of the author(s) and are not necessarily endorsed by the U.S. Army.

\bibliography{ITL_CAM_YOLO}

\end{document}